\documentclass[10pt,twocolumn,letterpaper]{article}

\usepackage[pagenumbers]{cvpr} 

\usepackage{tabularx}
\usepackage{booktabs}
\usepackage{array}
\usepackage[table,xcdraw]{xcolor}
\usepackage{float}

\definecolor{cvprblue}{rgb}{0.21,0.49,0.74}
\usepackage[pagebackref,breaklinks,colorlinks,allcolors=cvprblue]{hyperref}

\definecolor{oai-white}{HTML}{FFFFFF}
\definecolor{oai-black}{HTML}{000000}
\definecolor{oai-red}{HTML}{FF4500}
\definecolor{oai-green}{HTML}{51DA4C}
\definecolor{oai-blue}{HTML}{0000FF}
\definecolor{oai-yellow}{HTML}{FFF639}
\definecolor{oai-magenta}{HTML}{FF45FF}
\definecolor{oai-cyan}{HTML}{00FFFF}
\definecolor{oai-orange}{HTML}{FE7600}
\definecolor{oai-violet}{HTML}{8A2BE2}
\definecolor{oai-brown}{HTML}{A0522D}
\definecolor{oai-green-050}{HTML}{F4FFF4}
\definecolor{oai-green-100}{HTML}{E9FFE8}
\definecolor{oai-green-200}{HTML}{D9FFD8}
\definecolor{oai-green-300}{HTML}{C9FFC7}
\definecolor{oai-green-400}{HTML}{A6FFA3}
\definecolor{oai-green-500}{HTML}{7CF178}
\definecolor{oai-green-600}{HTML}{51DA4C}
\definecolor{oai-green-700}{HTML}{3FA93B}
\definecolor{oai-green-800}{HTML}{2D712A}
\definecolor{oai-green-900}{HTML}{193718}
\definecolor{oai-gray-000}{HTML}{FFFFFF}
\definecolor{oai-gray-100}{HTML}{FAFAFA}
\definecolor{oai-gray-200}{HTML}{F5F5F5}
\definecolor{oai-gray-300}{HTML}{E5E5E5}
\definecolor{oai-gray-400}{HTML}{FFB7A4}
\definecolor{oai-gray-500}{HTML}{CDCDCD}
\definecolor{oai-gray-600}{HTML}{A8A8A8}
\definecolor{oai-gray-700}{HTML}{747474}
\definecolor{oai-gray-800}{HTML}{393939}
\definecolor{oai-gray-900}{HTML}{000000}
\definecolor{visual}{HTML}{A50E0E}       
\definecolor{linguistic}{HTML}{174EA6}   
\definecolor{relational}{HTML}{E37400}   
\definecolor{egocentric}{HTML}{0D652D}

\usepackage{graphicx}
\usepackage{amssymb}
\usepackage{adjustbox}
\usepackage{colortbl}
\usepackage{xcolor}
\usepackage{multirow}
\usepackage{array}
\usepackage{makecell}
\usepackage{bbm}
\usepackage{collcell,xfp}
\usepackage{pgf}
\usepackage{tikz}
\usepackage[most]{tcolorbox}
\usepackage{csquotes}
\usepackage[noorphans,vskip=1em,leftmargin=1em]{quoting}
\usepackage{pifont}
\usepackage{enumitem} 
\usepackage{tcolorbox} 
\usepackage{forest}
\usepackage{wrapfig}
\usepackage{caption}
\usepackage{subcaption}
\usepackage{longtable}
\usepackage[T1]{fontenc}

\colorlet{mapcolor}{ForestGreen}

\title{Can World Simulators Reason? \\Gen-ViRe: A Generative Visual Reasoning Benchmark}

\author{%
  \textbf{Xinxin Liu$^{1}$\thanks{Equal contribution.}~~, Zhaopan Xu$^{2}$\footnotemark[1]~~, \textbf{Ming Li$^{1}$}, \textbf{Kai Wang$^{2}$}, \textbf{Yong Jae Lee$^{3}$}, \textbf{Yuzhang Shang$^{1}$}\thanks{Corresponding author}}\\
  $^{1}$University of Central Florida,
  $^{2}$National University of Singapore, $^{3}$UW-Madison \\
\href{https://l-codingspace.github.io/gvr_web}{\textbf{Project}}~~~~~~\href{https://github.com/L-CodingSpace/GVR}{\textbf{Code}}
}

\begin{document}
\maketitle
\begin{abstract}
While Chain-of-Thought (CoT) prompting enables sophisticated symbolic reasoning in LLMs, it remains confined to discrete text and cannot simulate the continuous, physics-governed dynamics of the real world. Recent video generation models have emerged as potential world simulators through Chain-of-Frames (CoF) reasoning---materializing thought as frame-by-frame visual sequences, with each frame representing a physically-grounded reasoning step. Despite compelling demonstrations, a challenge persists: existing benchmarks, focusing on fidelity or alignment, do not assess CoF reasoning and thus cannot measure core cognitive abilities in multi-step planning, algorithmic logic, or abstract pattern extrapolation. This evaluation void prevents systematic understanding of model capabilities and principled guidance for improvement. We introduce Gen-ViRe (Generative Visual Reasoning Benchmark), a framework grounded in cognitive science and real-world AI applications, which decomposes CoF reasoning into six cognitive dimensions --from perceptual logic to abstract planning --and 24 subtasks. Through multi-source data curation, minimal prompting protocols, and hybrid VLM-assisted evaluation with detailed criteria, Gen-ViRe delivers the first quantitative assessment of video models as reasoners. Our experiments on SOTA systems reveal substantial discrepancies between impressive visual quality and actual reasoning depth, establishing baselines and diagnostic tools to advance genuine world simulators.
\end{abstract}

\begin{figure}[t]
    \centering
    \includegraphics[width=1.0\linewidth]{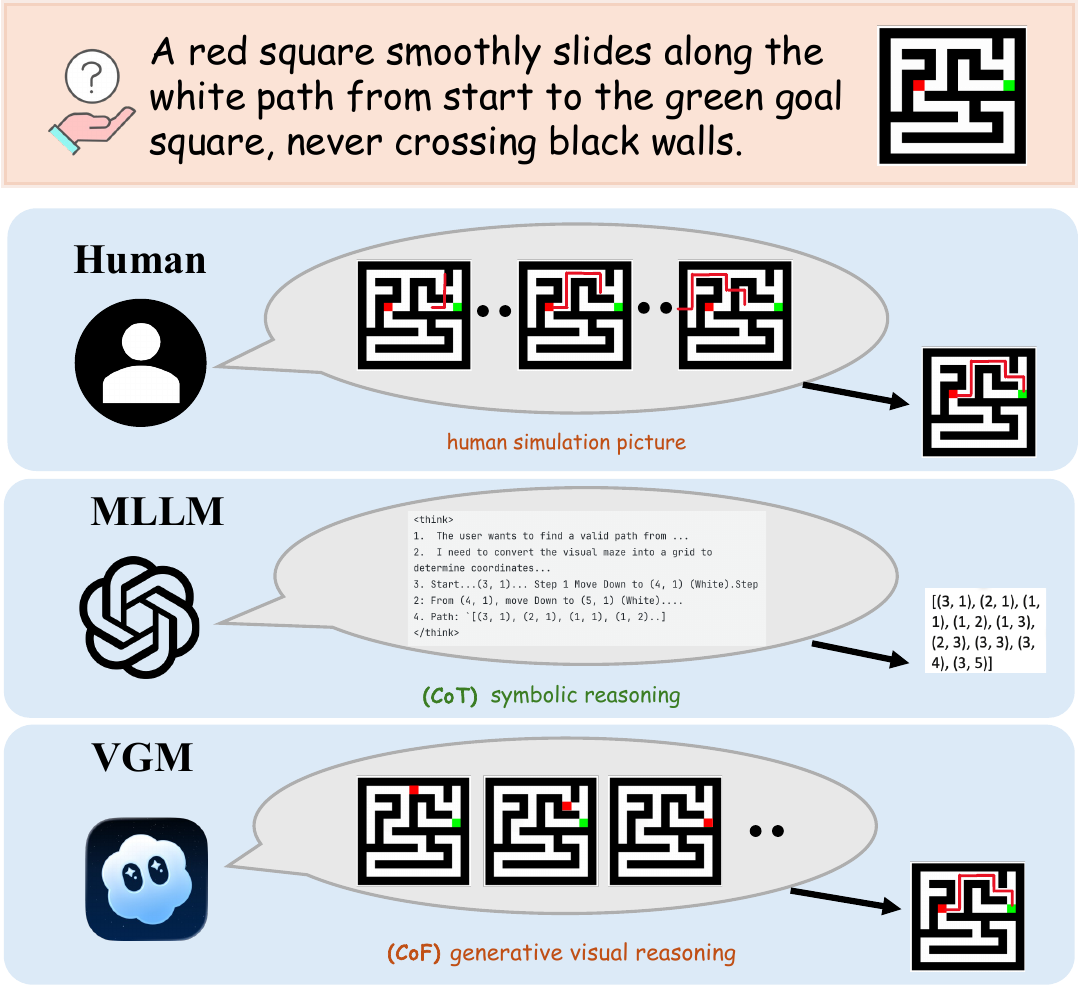}
    \vspace{-0.5cm}
    \caption{A comparison of reasoning approaches for a maze-solving task. Humans visualize the path via mental simulation. A Multimodal Large Language Model (MLLM) uses symbolic reasoning (CoT) to describe the path, e.g., via coordinates. In contrast, a Video Generation Model (VGM) uses generative visual reasoning (CoF) to physically simulate the process, generating frames of the square moving from start to finish.}
    \label{fig:maze}
    \vspace{-0.4cm}
\end{figure}

\section{Introduction}

Recent breakthroughs in Chain-of-Thought (CoT) prompting have demonstrated that large-scale language models (LLMs) can exhibit emergent reasoning capabilities across diverse symbolic domains, from mathematical problem-solving to code generation~\cite{cot, gpt-o1, deepseek-r1}. These advances represent a significant milestone in AI's ability to perform complex reasoning. However, CoT operates exclusively in the symbolic realm—reasoning over language tokens about abstract concepts. While this enables powerful logical inference, it fundamentally cannot capture the continuous, visual, and spatial nature of the real world. This limitation is evident in tasks ranging from abstract spatial navigation to complex physical interaction. 

For instance, in the maze-solving task (shown in Figure~\ref{fig:maze}), when an MLLM using CoT is asked to find the path, it describes the solution symbolically—perhaps by outputting a list of coordinates [e.g., "(3, 1), (2, 1)..."]. It cannot, however, simulate the actual, continuous, dynamic process of the "red square" moving through the path. Similarly, consider a robotic manipulation task: planning how to open a nailed wooden box. CoT can describe the steps (`use a crowbar, apply leverage'), but it cannot verify whether the plan is physically feasible—whether the robot's gripper can grasp the crowbar, whether the force angle is sufficient, or whether nearby objects will interfere. This reveals a critical limitation: \textit{to build AI systems that truly understand and simulate spatial dynamics and physical interaction, we need models that can reason through continuous visual simulation, not just symbolic description.}

This gap—the need for dynamic, generative reasoning—is also precisely where conventional computer vision paradigms fall short. Conventional vision models function as specialized, passive perceivers—excelling at singular discriminative tasks like object detection or segmentation, yet lacking the unified generative reasoning capabilities needed to simulate dynamic processes. Recognizing this gap, recent pioneering works have proposed a revolutionary paradigm shift: video models as zero-shot learners and world simulators~\cite{sora,genie,veo,cof}. As world simulators, these models are transitioning from merely rendering pixels to building an implicit spatiotemporal and physical engine by training on massive video data. This emergent ``simulator'' then enables them to act as zero-shot learners; they are no longer confined to the specific compositions seen during training but can, much like humans, leverage a fundamental understanding of the world to ``zero-shot'' imagine, reason about, and create entirely new, logical scenarios.
At the core of this new paradigm lies the \textbf{Chain-of-Frames (CoF)} mechanism~\cite{cof}, where reasoning is actualized through frame-by-frame video generation. Unlike CoT's discrete symbolic transitions (text $\rightarrow$ text), CoF materializes reasoning as a goal-directed, continuous visual state evolution (frame $\rightarrow$ frame). When a model solves a visual Sudoku puzzle, navigates a maze, or plans multi-step tool use through CoF, each generated frame represents an incremental reasoning step that is both physically grounded and temporally coherent. The generative process itself becomes an act of thinking, not just describing the symbolic coordinates for the maze but generating the dynamic trajectory of the red square moving. It transforms models from passive descriptors into executable world simulators.

The potential of this paradigm is evidenced by rapid progress in video generation technology. Large-scale models---from commercial systems like Sora~\cite{sora} and Kling~\cite{kling} to open models like CogVideo~\cite{cogvideo} and HunyuanVideo~\cite{hunyuanvideo}---have demonstrated impressive emergent capabilities in understanding physical interactions, temporal causality, and spatial relationships. Qualitative demonstrations from pioneering works are compelling: models generating coherent multi-step processes, respecting object permanence, and exhibiting intuitive physics. Yet despite this progress, a fundamental question remains unanswered: \textit{how well do these models actually reason?} Qualitative showcases, while impressive, provide no systematic measure of reasoning depth. More critically, existing video generation benchmarks, which focus primarily on fidelity, alignment, or object-level consistency, do not assess CoF reasoning and therefore cannot measure a model's core cognitive abilities in multi-step planning and algorithmic logic, or abstract pattern extrapolation~\cite{huang2024vbench, liu2024evalcrafter, liu2023fetv, fan2024aigcbench, han2025videobench, sun2025t2vcompbench, kou2024subjective, doersch2022tapvid}. Without rigorous quantitative evaluation, we cannot distinguish genuine reasoning from sophisticated pattern matching, nor can we diagnose where models fail---perception, physics understanding, or planning---to guide systematic improvement.

\begin{figure}[t!] 
    \centering 
    \centering
    \includegraphics[width=0.9\linewidth]{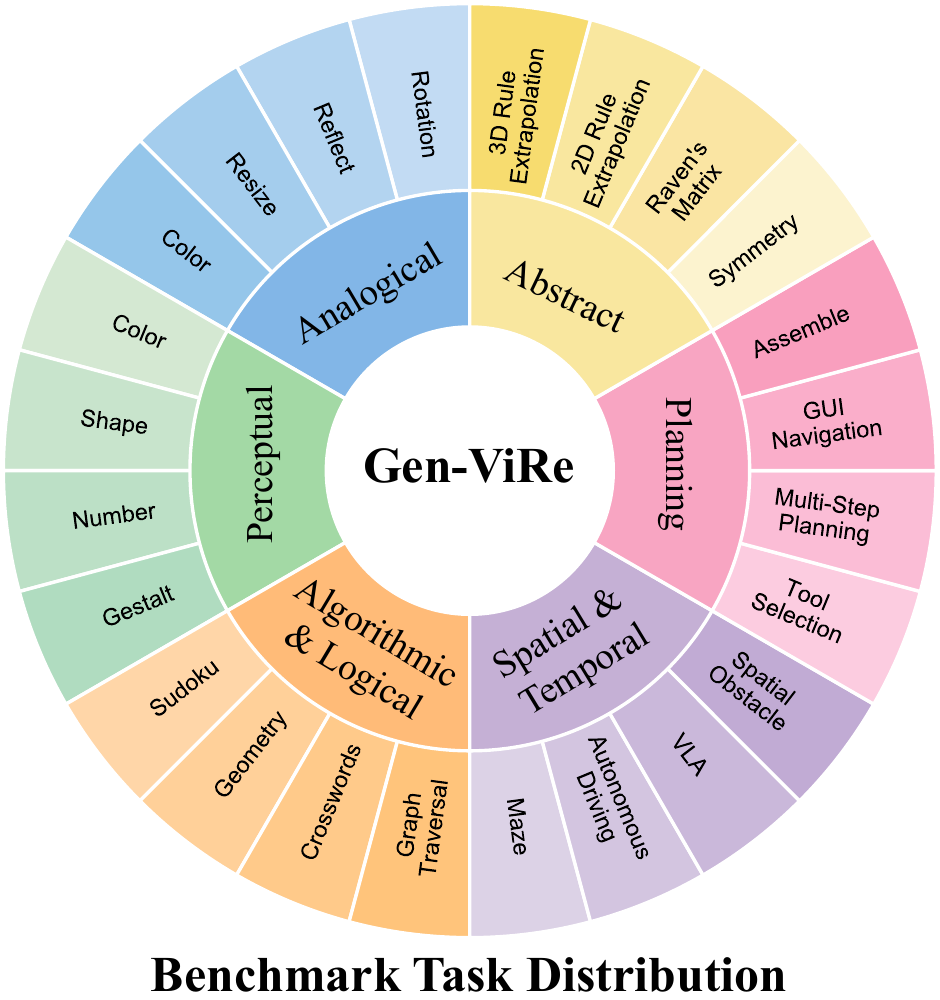} 
    \vspace{-0.3cm}
    \caption{Our Gen-ViRe evaluates six core cognitive dimensions: (1) Perceptual, (2) Analogical, (3) Abstract, (4) Planning, (5) Spatial \& Temporal, and (6) Algorithmic \& Logical, with each dimension comprising four different sub-categories.}
    \label{fig:pie_and_radar}
    \vspace{-0.4cm}
\end{figure}

\begin{figure*}[t]
    \centering
    \includegraphics[width=\textwidth]{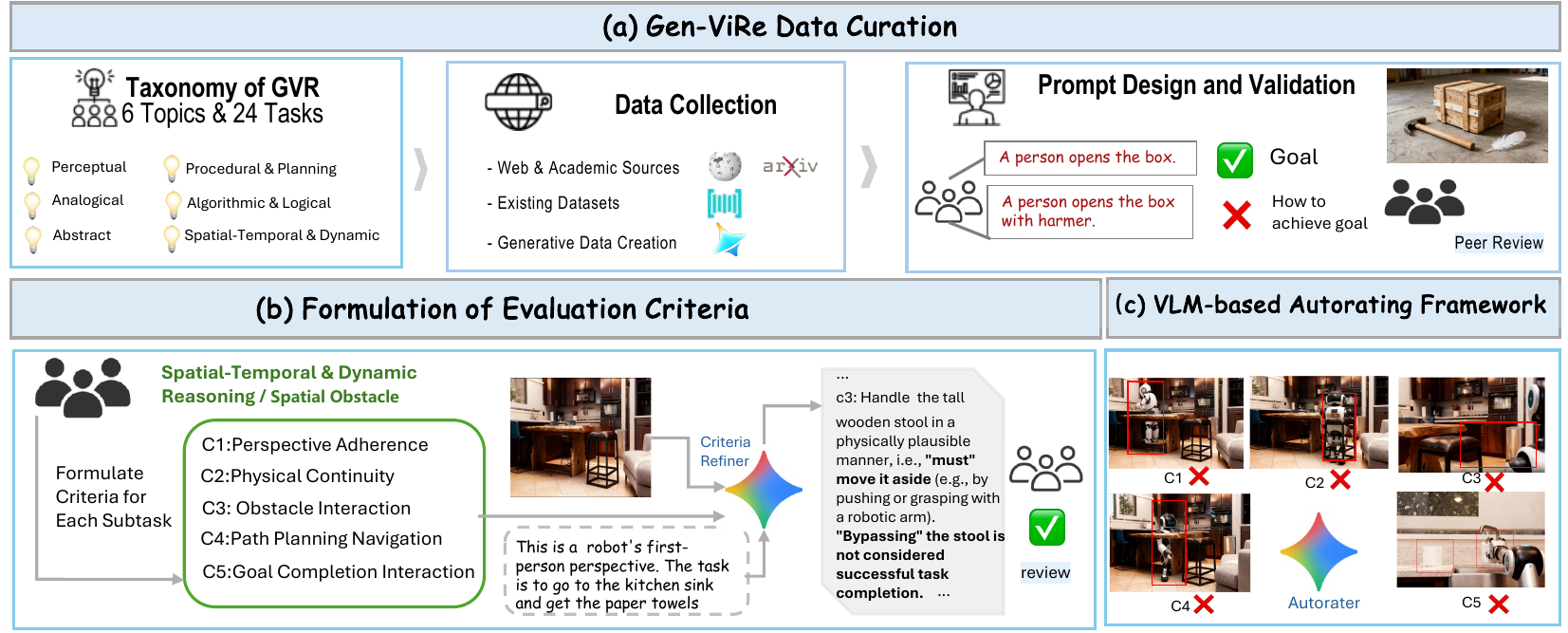}
    \vspace{-0.7cm}
    \caption{The evaluation framework of Gen-ViRe. \textbf{(a)} Data Curation: Shows the benchmark development process, including defining the taxonomy, collecting data from multiple channels (web, existing datasets, AI generation), and designing \& validating prompts through Peer Review. \textbf{(b)} Formulation of Evaluation Criteria: Demonstrates the process of formulating detailed, multi-dimensional evaluation criteria (as shown by C1-C5 in the figure) for each prompt of every subtask. \textbf{(c)} VLM-based Autorating Framework: Illustrates how the VLM (Autorater) conducts item-by-item analysis and automatic scoring of the generated videos based on the specific criteria defined in (b).}
    \vspace{-0.3cm}
    \label{fig:data_eval}
\end{figure*}

To address this critical evaluation gap, we propose \textbf{Gen-ViRe} (Generative VisualReasoning Benchmark), the first comprehensive framework purpose-built to systematically assess CoF reasoning by video generation models. Gen-ViRe is grounded in dual foundations: cognitive science principles defining core pillars of human reasoning
\cite{ kahneman2011thinking, gentner1983structuremapping, johnsonlaird1983mentalmodels, raven2000ravenreview, newell1972humanproblemsolving, byrne2007rationalimagination}
, and application-driven requirements from emerging domains like embodied AI and autonomous systems
\cite{feng2025embodied, Zhang2025EmbodiedVSR}.
We decompose Generative VisualReasoning into six fundamental dimensions—
Perceptual, Analogical, Algorithmic \& Logical, Spatio-Temporal \& Dynamic,
Procedural \& Planning, and Abstract Reasoning—spanning the complete spectrum
from foundational perception to high-order planning
\cite{BoberIrizar2024NeuralAbstraction}.
Our data curation employs a multi-source strategy as shown in Figure~\ref{fig:data_eval}: curating from web resources and academic papers, integrating existing datasets (e.g., ARC-AGI for abstract reasoning, KiVA for analogical tasks), and generating novel tasks using T2I models where suitable data is absent. We implement rigorous prompt design through minimal prompting principles—deliberately providing only high-level goals to assess autonomous reasoning rather than instruction-following—validated through iterative peer review. Our evaluation methodology features a hybrid VLM-assisted approach: Image VLMs judge final outputs for static reasoning tasks, while Video VLMs assess the complete generation process for dynamic tasks, with each judge provisioned with detailed, sub-category-specific criteria refined through multi-round human validation. Through extensive experiments evaluating 7 state-of-the-art video generation models across 72 distinct reasoning prompts (with each model generating 5 instances per prompt, totaling 360 videos per model and over 2,500 overall), Gen-ViRe delivers the first systematic quantification of CoF reasoning capabilities, revealing critical gaps between qualitative potential and quantitative performance, and establishing scientific baselines to guide the development of genuine world simulators.

Our main contributions are as follows:
\begin{itemize}
    \item We propose \textbf{Gen-ViRe}, the first comprehensive benchmark specifically designed to systematically evaluate Chain-of-Frames reasoning across six fundamental dimensions grounded in cognitive science and practical AI applications, providing rigorous scientific assessment from foundational perception to high-order planning.
    
    \item We establish a complete \textbf{evaluation methodology} combining minimal prompting design, multi-source data curation (web/academic resources, existing datasets, generative creation), and hybrid VLM-assisted assessment with detailed criteria, enabling quantitative measurement of generative reasoning and failure mode diagnosis.
    
    \item Through extensive experiments on state-of-the-art video generation models, we provide the \textbf{first systematic analysis} of current CoF reasoning capabilities, revealing significant gaps between qualitative demonstrations and quantitative performance, and establishing baselines for future research toward genuine world simulators.
\end{itemize}
\section{Related Work}

\subsection{\textbf{Discrete Chain-of-Thought Reasoning.}}
\vspace{-0.1cm}
Chain-of-Thought (CoT) prompting has revolutionized how large language models approach complex reasoning tasks by explicitly generating intermediate reasoning steps~\cite{cot}. GPT-o1~\cite{gpt-o1} demonstrated that LLMs can leverage CoT for test-time scaling—trading computation for reasoning depth—while DeepSeek-R1~\cite{deepseek-r1} advanced this through Reinforcement Learning with Verifiable Rewards (RLVR), democratizing sophisticated reasoning via open-source release. Recent work has extended CoT to multimodal domains: GPT-4o~\cite{gpt-4o} pioneered vision-language joint reasoning, and Bagel~\cite{bagel} leverages vision-text interleaved CoT to enhance visual generation. However, CoT remains fundamentally symbolic and discrete—operating in the space of language tokens rather than continuous visual states. While effective for formal logic, CoT is inherently discriminative and passive: it cannot dynamically simulate the physical evolution of the visual world. A CoT chain can describe "a ball rolls down a slope," but cannot generate the actual frame-by-frame trajectory governed by gravity and collision. This limitation necessitates a new paradigm grounded in continuous, generative visual simulation.

\subsection{\textbf{Continuous Chain-of-Frames Reasoning.}}
\vspace{-0.1cm}
The emerging Chain-of-Frames (CoF) paradigm represents a revolutionary shift in how AI systems perform reasoning~\cite{cof}. Unlike CoT, which operates on discrete symbolic transitions (text → text), CoF materializes reasoning as continuous visual state evolution (frame → frame). In this paradigm, reasoning is not merely described—it is executed and visualized through frame-by-frame video generation. When a model generates a sequence showing how to solve a visual puzzle, navigate a maze, or manipulate objects to achieve a goal, each generated frame represents an incremental reasoning step that is physically grounded and temporally coherent.
Recent pioneering works have begun to explore this new frontier. The Genie series~\cite{genie} demonstrates how interactive world generation can enable embodied reasoning, allowing agents to learn causal relationships through simulated physical interactions. Veo-3 pushes the boundaries of long-horizon reasoning, generating extended video sequences that maintain spatial-temporal consistency across complex dynamic scenes~\cite{veo}. These works provide compelling qualitative evidence that video models have the potential as world simulators, capable of understanding physics, causality, and spatial relationships~\cite{cof}.

\subsection{\textbf{Video Generation: Potential World Simulators.}}
\vspace{-0.1cm}
Recent years have witnessed remarkable progress in video generation, predominantly driven by diffusion-based architectures. Models such as Sora~\cite{sora}, Kling~\cite{kling}, and SeeDance~\cite{seedance} have achieved breakthroughs in visual quality, generation length, and resolution through large-scale diffusion transformers. While these commercial models remain closed-source, the research community has also contributed open-sourced models like CogVideo~\cite{cogvideo} and HunyuanVideo~\cite{hunyuanvideo} and WanVideo~\cite{wanvideo}. These models demonstrate impressive implicit learning of physical principles purely from observing massive video datasets. Crucially, these models are evolving beyond mere visual generators—they are becoming potential world simulators. For instance, when prompted to generate a video of ``a basketball bouncing downstairs'', Veo-3~\cite{veo} can produce sequences that respect gravity, conserve momentum, and exhibit realistic deformation upon impact~\cite{cof}.

\section{Gen-ViRe Benchmark}
\begin{figure*}[t]
    \centering
    \includegraphics[width=\textwidth]{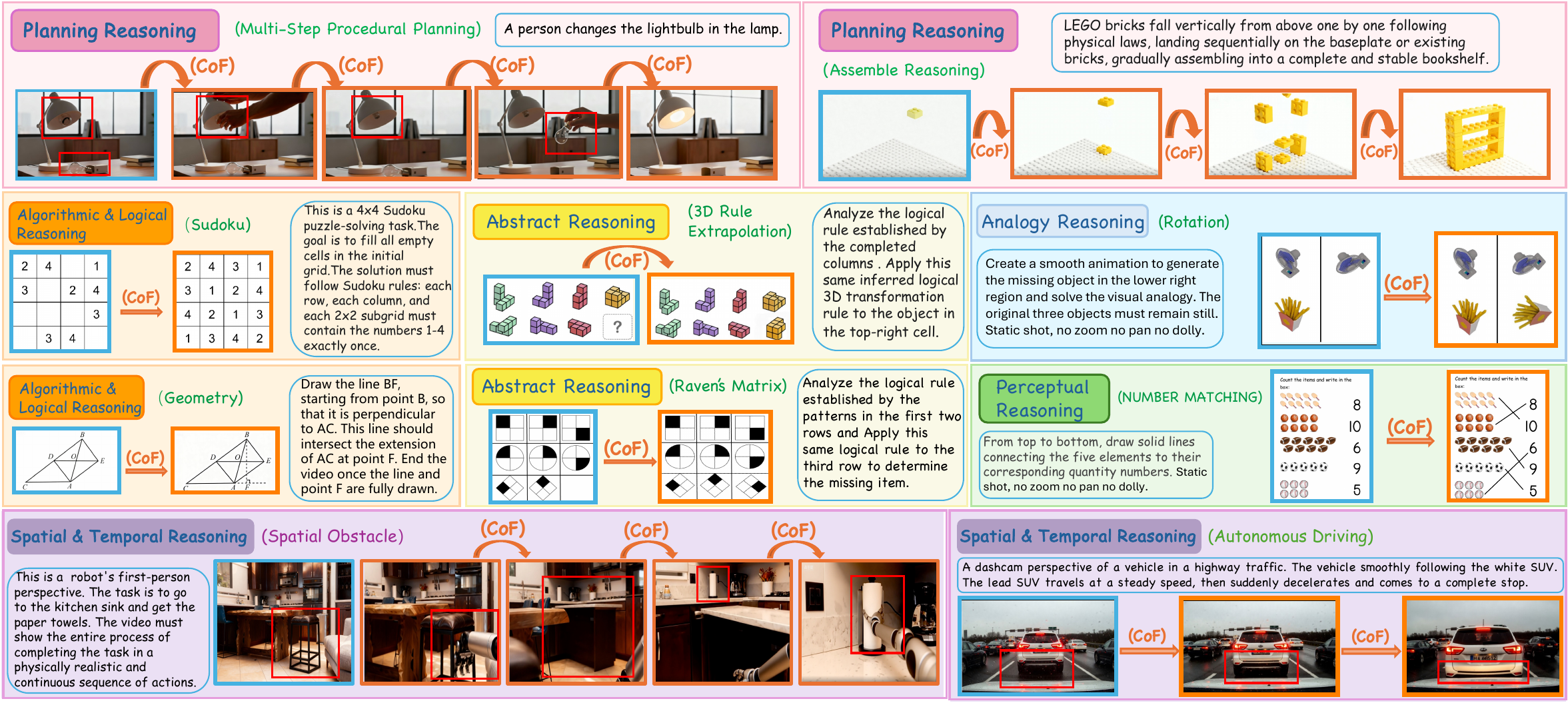}
    \vspace{-0.7cm}
    \caption{Qualitative examples of Gen-ViRe tasks. It illustrates sample inputs and their expected Chain-of-Frames (CoF) visual reasoning outputs across the six cognitive dimensions, highlighting the benchmark's breadth from foundational perception to high-order planning.}
    \vspace{-0.3cm}
    \label{fig:main}
\end{figure*}

\subsection{Definition: Generative Visual Reasoning}
\label{sec:problem_definition}

In this paper, we aim to evaluate an emergent capability beyond traditional discriminative VQA, which we term \textbf{Generative Visual Reasoning (GVR)}. We conceptually define GVR as an agent's ability to solve a complex visual problem by simulating a spatio-temporal dynamic or executing a multi-step plan through a sequential, generative process. Unlike traditional benchmarks that aim to evaluate a single, deterministic final answer (e.g., a class label or a bounding box), the goal of a GVR task is to evaluate the \textbf{generation process} (i.e., a frame sequence $F$) itself for its logical coherence, physical plausibility, and goal-orientation. This Generative Visual Reasoning (GVR) capability is primarily actualized through the \textbf{``Chain-of-Frames" (CoF)} mechanism. We can formalize this generative process as an autoregressive sequence:

\begin{equation}
F = \{f_1, f_2, \dots, f_N\}, \space f_i = g(I_i, q, F_{<i}),
\end{equation}
where $q$ is the initial context or task query (e.g., ``Solve this maze''), $F$ is the complete reasoning sequence, and $F_{<i} = \{f_1, \dots, f_{i-1}\}$ is the generation history up to the previous step. $f_i$ is the $i$-th step in the sequence. As this work focuses on video generation models, $f_i$ represents the next generated video frame. The entire reasoning sequence $F$ thus constitutes a complete and coherent video, which itself is the manifestation of the model's ``thinking" process.


\label{taxonomy}
\subsection{Taxonomy of Generative Visual Reasoning}
\label{sec:taxonomy}

Our proposed Gen-ViRe taxonomy is rooted in two complementary perspectives: (i) the theoretical foundations of cognitive science and (ii) the practical demands of key emerging GVR application domains. More specifically: \noindent{\textbf{(i) Cognitive Science Foundations}:} A robust GVR benchmark should be able to evaluate the core pillars of human cognitive ability~\cite{ARC-AGI,BoberIrizar2024NeuralAbstraction,byrne2007rationalimagination,gentner1983structuremapping,johnsonlaird1983mentalmodels,kahneman2011thinking,newell1972humanproblemsolving},. This provides the theoretical basis for our categories of Perceptual Reasoning, Analogical Reasoning, and Abstract Reasoning. \noindent{\textbf{(ii) Application-Driven Requirements}:} Concurrently, as AI expands from the symbolic to the physical realm, emerging applications (e.g., Embodied AI, Autonomous Driving) demand dynamic, multi-step reasoning~\cite{gu2024controlvla}. These capabilities are essential for the next frontier of AI but cannot be evaluated by existing benchmarks. These practical requirements provide the basis for our categories of Planning Reasoning, Spatial/Temporal Reasoning, and Algorithmic \& Logical Reasoning.

Grounded on the above two pillars, we decompose GVR into six critical and complementary dimensions that form a full spectrum of capabilities, from foundational perception to high-order planning: (1) Perceptual Reasoning; (2) Analogical Reasoning; (3) Algorithmic \& Logical Reasoning; (4) Spatial\&Temporal \& Dynamic Reasoning; (5) Procedural \& Planning Reasoning; (6) Abstract Reasoning. This framework is not only theoretically robust but also practically essential. It provides the first comprehensive evaluation framework for this new paradigm, allowing us to scientifically quantify and diagnose the nascent reasoning abilities driven by the Chain-of-Frames paradigm.

\noindent{\textbf{Perceptual Reasoning}.}
This category probes a model's foundational cognition: the ability to move beyond passive perception and actively reason about the logical relationships between visual attributes. We test four key logic types via worksheet-style puzzles: association (color), morphology (shape), quantification (quantity-to-numeral), and Gestalt (part-to-whole). Models must execute a precise, procedural, spatio-temporal action (e.g., ``draw a connecting line") to demonstrate their conclusion, rather than just outputting a static answer.

\noindent{\textbf{Spatial \& Temporal Reasoning}.}
Tasks in this category assess a model's ability to reason about motion, causality, and change within a continuous scenarios. These tasks require the model to generate a temporally coherent and physically plausible chain-of-frames (CoF). We probe faculties such as the ability to predict and model motion and to plan and execute navigation under constraints (e.g., Autonomous Driving, VLA Manipulation, Maze Traversal, Spatial Obstacle Navigation). It measures the model's ability to build an internal world model.

\noindent{\textbf{Planning Reasoning}.}
This category targets a model's higher-order cognitive ability to perform multi-step, goal-directed planning. It requires models to decompose a complex goal into a discrete, logical, and correctly-ordered sequence of sub-actions. We probe four domains: (1) Causal Tool Reasoning (e.g., selecting the correct tool); (2) Sequential Task Decomposition (e.g., changing a lightbulb); (3) Hierarchical Digital Planning (e.g., GUI navigation); and (4) Physically-Constrained Assembly (e.g., brick-by-brick construction with physical stability).

\noindent{\textbf{Analogical Reasoning}.}
This dimension focuses on the capability of relational abstraction. We adopt the classic visual analogy task ($A \rightarrow B$ :: $C \rightarrow ?$). The model must perform a two-stage inferential process: (1) Discover the latent transformation rule by comparing the source pair $(A \rightarrow B)$; and (2) Apply this inferred rule to the new target object $C$.

\noindent{\textbf{Algorithmic \& Logical Reasoning}.}
Here, we evaluate a model's ability to follow and execute formal rules and constraints. It requires the model to apply symbolic reasoning to the visual domain to solve intellectual puzzles, including Visual Sudoku, Graph Traversal, Geometry, and Crosswords. In these tasks, the model must demonstrate its understanding of abstract rules and successfully apply them within the visual context to arrive at a correct solution.

\noindent{\textbf{Abstract Reasoning}.}
This category measures a model's highest-order cognitive ability: identifying and extrapolating abstract patterns and rules. This is closely related to human ``fluid intelligence." Tasks require the model to look beyond superficial features to discover the hidden generative principles in the data. Tasks we test include Symmetry, 2D/3D Rule Extrapolation, and Raven's Progressive Matrices. Success in these tasks indicates the model is not just a pattern mimic but a rule discoverer.

\section{Data Curation}
\subsection{Data Collection}

As described in Section~\ref{sec:taxonomy}, our taxonomy covers six key dimensions. To construct a diverse and information-rich benchmark capable of comprehensively evaluating GVR abilities, we designed a multi-source data collection strategy tailored to the unique requirements of each category. Our data sources are primarily divided into three categories: Web and Academic Sources, Integration of Existing Datasets, and Generative Data Creation.

\noindent{\textbf{Web and Academic Sources.}} The foundation of our data collection comes from public web resources and academic publications. We use targeted keywords to collect candidate images from search engines like Google. Concurrently, we extract high-quality figures, illustrations, and qualitative examples from relevant academic papers~\cite{gu2024controlvla}~\cite{yang2025mindjourney}. Notably, for the Perceptual Reasoning category, we also sourced numerous children's intelligence tests from the web (e.g., color, shape, and quantity matching puzzles). All collected materials underwent rigorous manual screening and secondary editing (e.g., cropping, annotation, content modification)---such as the modifications made to the aforementioned Spatial Obstacle image---to ensure they perfectly align with the logical and visual constraints of our benchmark tasks.

\noindent{\textbf{Integration of Existing Datasets}.} To evaluate model capabilities in specific domains, we extracted or adapted tasks from several public datasets. This includes GUI navigation datasets~\cite{li2025screenspot} for Planning Reasoning, Geometry datasets~\cite{geolaux} for Algorithmic \& Logical Reasoning, and tasks borrowed from KiVA~\cite{kiva} for Analogical Reasoning. Critically, to test high-order abstract reasoning, we also incorporated challenging tasks from the ARC-AGI benchmark~\cite{ARC-AGI}, which is widely considered a gold standard for evaluating fluid intelligence.

\noindent{\textbf{Generative Data Creation}.} For many tasks in our benchmark---especially in the Planning Reasoning category (e.g., Tool Selection and Use)---no large-scale, logically-consistent datasets readily exist. Inspired by the qualitative examples in the pioneering work on Chain-of-Frames~\cite{cof}, we defined the generative rules and underlying logic for these advanced reasoning tasks and leveraged advanced Text-to-Image models to create entirely new visual puzzles. This approach allows us to systematically control task difficulty, compositional complexity, and generalization requirements, which is unachievable by passively collecting existing data.

\subsection{Prompt Design and Validation}

\textbf{Prompt Design and Validation.} Our prompt design and validation process ensures high task fidelity through a minimal prompting principle and a rigorous, iterative peer-review workflow. It begins with a core design adherence to minimal prompting, aiming to assess the model's autonomous reasoning capabilities rather than its proficiency in adhering to complex instructional formats. For instance, in our Embodied Spatial Obstacle task, the model is provided with an initial state image (e.g., a first-person robot view of a kitchen, blocked by a large table and a high stool) and a high-level objective (e.g., ``This is a robot's first-person perspective. The task is to go to the coffee machine in the kitchen and get the paper towels.''). We deliberately abstain from specifying \textit{how} to handle the obstacles. A successful output requires the model to autonomously reason about the implicit physical and spatial constraints. The prompt does not mention the obstacles, but the model must infer that (1) it must navigate \textit{around} the large table, not pass through it, (2) it must perform the \textit{mandatory} action of physically \textit{moving the stool aside}, as bypassing or stepping over it is defined as a failure, and (3) the interaction must be performed by a \textit{robotic manipulator}, not a human hand, to adhere to the ``robot's perspective'' constraint.

\noindent{\textbf{Iterative Peer Review Process}.}
To ensure the clarity and robustness of all task prompts, we implemented a strict, iterative peer-review process. A task draft formulated by one annotator is submitted to at least one other independent annotator for review. This review scrutinizes the task for clarity, potential ambiguities, and whether the ground truth is an indisputable, sole answer. Any flagged issues are returned for team discussion and revision. A primary focus of this process is the resolution of ``ambiguous references,'' a common source of model error. Our annotation team is trained to identify and rectify such vague language (e.g., ambiguous pronouns like ``it'' or ``that''), replacing them with precise descriptions to ensure high prompt fidelity.

\section{Evaluation Methodology}
\vspace{-0.2cm}
\noindent{\textbf{Formulation of Evaluation Criteria}.}
The core of our evaluation pipeline is the development of detailed criteria for each task subcategory. This formulation process is a unique, hybrid approach combining VLM assistance with multi-round human refinement. First, our team drafts preliminary evaluation standards for each task. We then provide these preliminary standards, along with the corresponding input image, text prompt, and task objective, to Gemini 2.5 Pro. The model's role is to refine these standards into a more detailed, rigorous, and operational evaluation rubric based on the full context of the specific task. Finally, the detailed criteria generated with VLM assistance undergo a final multi-person review and refinement by our team to ensure absolute accuracy and consistency.

\noindent\textbf{VLM-Assisted Evaluation Methods.}
~\label{sec:VLM}
Our evaluation methodology employs powerful Vision Language Models (VLMs) as automated judges. We utilize Gemini 2.5 Pro \cite{Comanici2025Gemini} as our unified VLM judge, leveraging its respective modality capabilities based on the task's requirements. For tasks where the evaluation is contingent upon the final visual output (such as analogical or geometric reasoning), we employ Gemini 2.5 Pro as an Image VLM judge. For more complex dynamic tasks (such as planning and spatial-temporal reasoning) that require assessing the entire generated process, we utilize Gemini 2.5 Pro as a Video VLM judge. Crucially, for every task, the designated VLM judge (Gemini 2.5 Pro) is provisioned with a detailed, sub-category-specific set of criteria. The judge then decomposes and assesses the model's output against each criterion, providing an independent score for its decision. This criteria-centric approach ensures a consistent and rigorous evaluation across the entire benchmark.


\section{Experiments}
\begin{table*}[!t]
\centering
\small
\setlength{\tabcolsep}{1pt}
\caption{Performance comparison across different reasoning categories for various video generation models. We highlight the top-three performing models in each column with varying shades of purple, where a darker shade indicates a higher rank.}
\vspace{-0.3cm}
\label{tab:main_table}
\renewcommand{\tabularxcolumn}[1]{m{#1}}
\begin{tabularx}{\textwidth}{@{}l
    >{\centering\arraybackslash}X
    >{\centering\arraybackslash}X
    >{\centering\arraybackslash}X
    >{\centering\arraybackslash}X
    >{\centering\arraybackslash}X
    >{\centering\arraybackslash}X
    >{\centering\arraybackslash}X
    >{\centering\arraybackslash}X
    @{}}
\toprule
\textbf{Methods} & 
\textbf{\#Videos} & 
\textbf{Avg.} & 
\textbf{\scriptsize Abstract} & 
\textbf{\scriptsize \parbox{2cm}{\centering Algorithmic \\ \& Logical}} & 
\textbf{\scriptsize Analogy} & 
\textbf{\scriptsize Perceptual} & 
\textbf{\scriptsize Planning} & 
\textbf{\scriptsize Spatio-Temporal} \\
\midrule
Kling-v1~\cite{kling} & 360 & 0.198 & 0.071 & 0.057 & 0.117 & 0.140 & 0.443 & 0.359 \\
Seedance-1.0-Lite~\cite{seedance} & 360 & 0.279 & 0.087 & 0.256 & 0.083 & 0.146 & 0.572 & 0.532 \\
Seedance-1.0-Pro~\cite{seedance} & 360 & 0.301 & 0.154 & 0.164 & 0.083 & 0.171 & 0.609 & \cellcolor[HTML]{9698ED}0.621 \\
Wan-2.5~\cite{wanvideo} & 360 & \cellcolor[HTML]{ECF4FF}0.490 & 0.412 & \cellcolor[HTML]{ECF4FF}0.411 & \cellcolor[HTML]{9698ED}0.500 & 0.378 & 0.702 & 0.536 \\
Veo-3.1~\cite{veo} & 360 & 0.486 & \cellcolor[HTML]{ECF4FF}0.440 & \cellcolor[HTML]{CBCEFB}0.451 & 0.367 & \cellcolor[HTML]{ECF4FF}0.386 & \cellcolor[HTML]{ECF4FF}0.722 & \cellcolor[HTML]{CBCEFB}0.550 \\
Hailuo-2.3~\cite{hailuo} & 360 & \cellcolor[HTML]{CBCEFB}0.493 & \cellcolor[HTML]{CBCEFB}0.494 & 0.355 & \cellcolor[HTML]{ECF4FF}0.383 & \cellcolor[HTML]{CBCEFB}0.425 & \cellcolor[HTML]{9698ED}0.778 & 0.524 \\
Sora-2~\cite{sora} & 360 & \cellcolor[HTML]{9698ED}0.560 & \cellcolor[HTML]{9698ED}0.604 & \cellcolor[HTML]{9698ED}0.472 & \cellcolor[HTML]{CBCEFB}0.483 & \cellcolor[HTML]{9698ED}0.496 & \cellcolor[HTML]{CBCEFB}0.768 & \cellcolor[HTML]{ECF4FF}0.537 \\
\bottomrule
\end{tabularx}
\end{table*}

\begin{figure*}[t]
    \centering
    \vspace{-0.1cm}
    \includegraphics[width=1.0\textwidth]{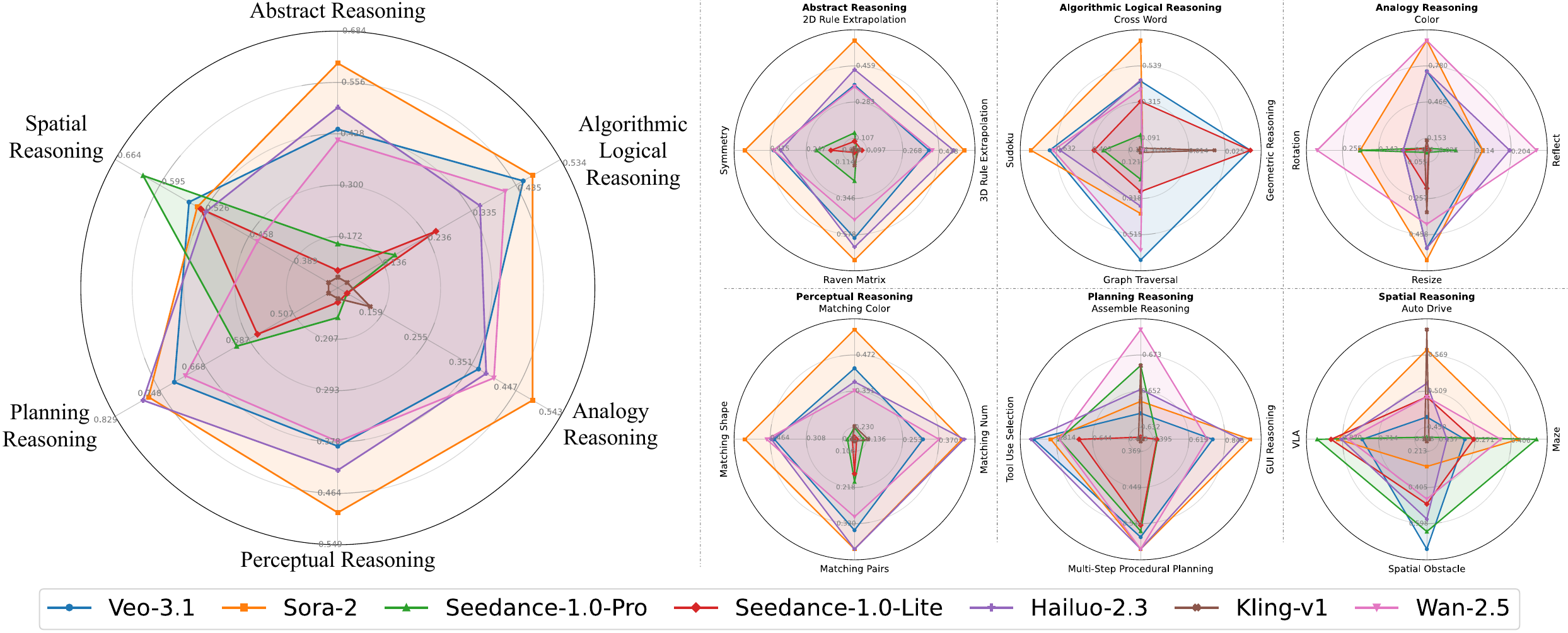}
    \vspace{-0.7cm}
    \caption{\textbf{Left:} The main chart compares the overall performance of the 7 state-of-the-art models across the six core cognitive dimensions (Abstract, Algorithmic, Analogy, Perceptual, Planning, and Spatial Reasoning). \textbf{Right:} The six sub-charts provide a detailed performance breakdown for the individual subtasks within each dimension. The legend (bottom) links each colored line to its respective model.}
    \label{fig:radar_main}
\end{figure*}
\label{sec:problem_definition}

\subsection{Experimental Setup}
\vspace{-0.1cm}
\noindent{\textbf{Evaluated Models.}}
\vspace{-0.1cm}
We evaluated a comprehensive suite of state-of-the-art (SOTA) video models, including Kling-v1, Seedance-1.0-Pro, Seedance-1.0-Lite, Veo-3.1, Sora-2, Wan-2.5, and Hailuo-2.3. Videos were generated in either 16:9 or 9:16 aspect ratio, determined by the native orientation of the input task image. For generation duration, we adopted the default 5-second for the Hailuo-2.3, 8-second for Veo-3.1 and Sora-2, and the default 10-second for Kling-v1, Seedance-1.0-Pro, Seedance-1.0-Lite, and Wan-2.5.

\noindent{\textbf{Evaluation Metrics.}}
The VLM-assisted evaluation process described in Section~\ref{sec:VLM} yields an output score for each task. To aggregate these individual scores, we follow the aggregation strategy of \textit{MEGA-BENCH} \cite{chen2024mega}. The score for each individual task is first normalized to a consistent [0, 1] range, where 1.0 signifies perfect adherence to all standards. Subsequently, to report comprehensive performance, we compute the macro-mean score of all normalized scores.


\subsection{Main Results}
\subsubsection{Qualitative Analysis and Case Studies}

\noindent\colorbox{lightgray}{\textbf{Case 1: Analogical Reasoning (Rotation and Color)}}
\begin{figure}[H]
    \centering
    \vspace{-0.4cm}
    \includegraphics[width=1.0\linewidth]{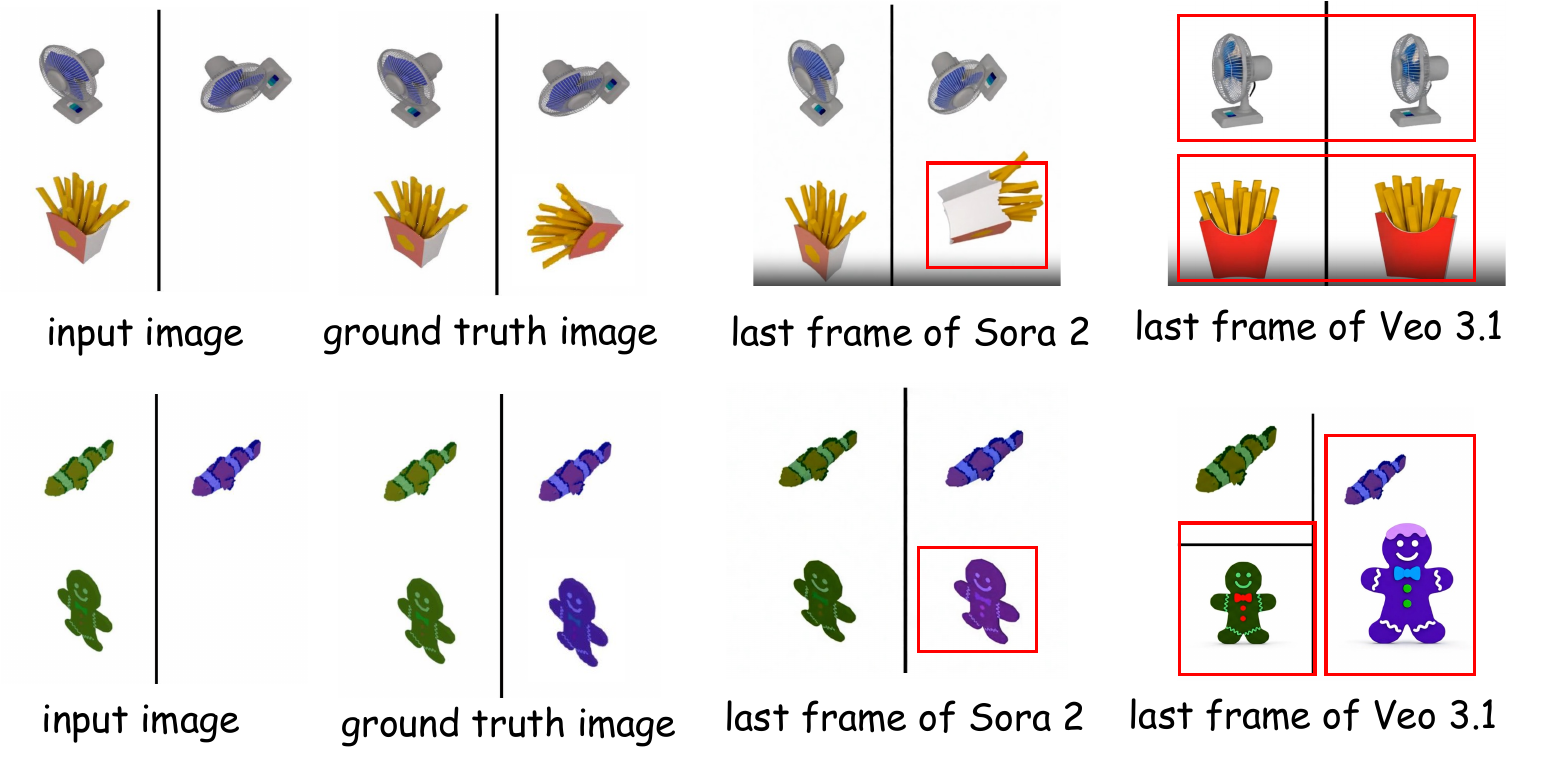}
    \vspace{-0.7cm}
    \caption{ \textbf{Showcase of Analogical Reasoning by Sora-2 and Veo-3.1.} On the complex Rotation task (top row), both Sora-2 and Veo-3.1 failed to recognize and apply the abstract rotational rule. In sharp contrast, both models easily solved the simpler Color analogy (bottom row), which only required matching an attribute.}
    \label{fig:model_pks_heatmap}
    \vspace{-0.4cm}
\end{figure}

\noindent{\textbf{I2V Prompt}}. Create a smooth animation to generate the missing object in the lower right region and solve the visual analogy. The original three objects must remain still. Static shot, no zoom no pan no dolly.

\noindent{\textbf{Takeaway}}. Current models' analogical reasoning performance correlates directly with task abstraction complexity: they can easily solve simple attribute matching (e.g., color) but expose core deficits when handling abstract, rule-based transformations (e.g., rotation).

\noindent\colorbox{lightgray}{\textbf{Case 2: Spatio-Temporal Reasoning (Spatial Obstacle)}}

\begin{figure}[H]
    \centering
    \vspace{-0.4cm}
    \includegraphics[width=\linewidth]{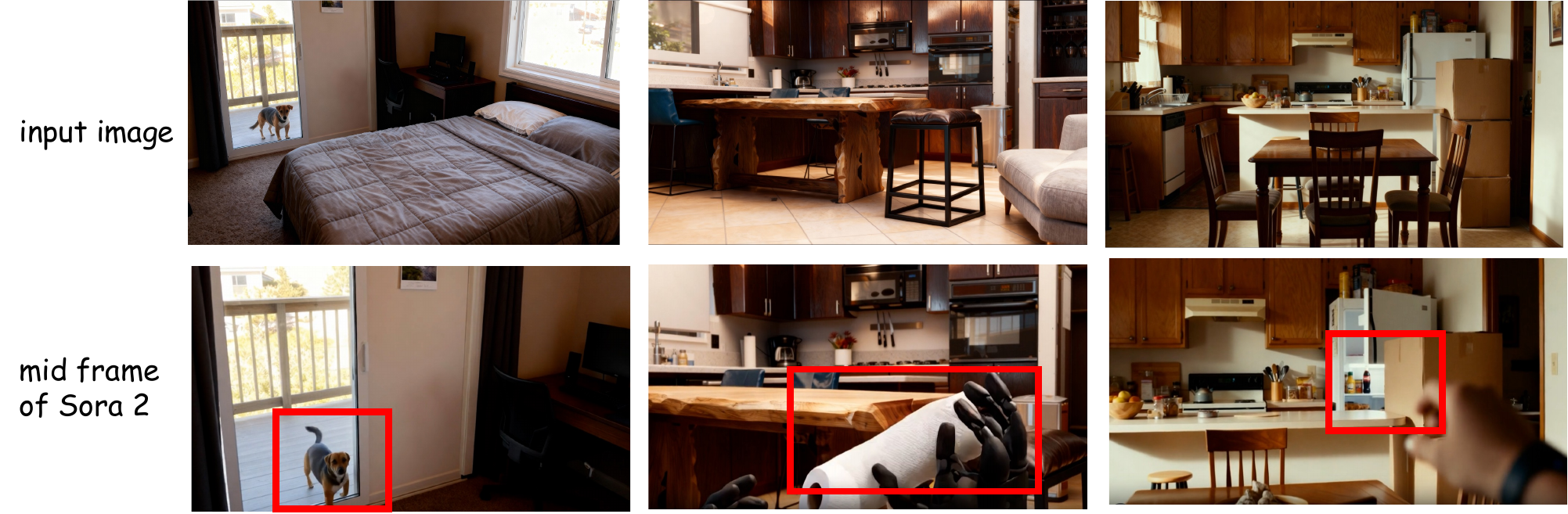}
    \vspace{-0.7cm}
    \caption{\textbf{Showcase of Sora-2's failures in Spatio-Temporal \& Dynamic Reasoning.} The top row shows input images; the bottom shows Sora-2's mid-frames. These frames reveal fundamental failures in simulating basic physical laws:  (Left) violating object permanence by showing a dog phasing through a closed glass door ; and (Middle) failing to simulate a continuous process, instead spawning paper towels into the scene ; (Right) depicting telekinesis, where an object is retrieved without contact.}
    \label{fig:model_pks_heatmap}
    \vspace{-0.4cm}
\end{figure}

\noindent{\textbf{Takeaway}.} Abstract logical reasoning (e.g., following algorithmic rules) and physical reality simulation (e.g., adhering to physical plausibility) are two distinct capabilities. The model's excellence in the former does not equate to mastery in the latter.

\noindent\colorbox{lightgray}{\textbf{Case 3: Algorithmic \& Logical Reasoning (Geometry)}}

\begin{figure}[H]
    \centering
    \vspace{-0.4cm}
    \includegraphics[width=\linewidth]{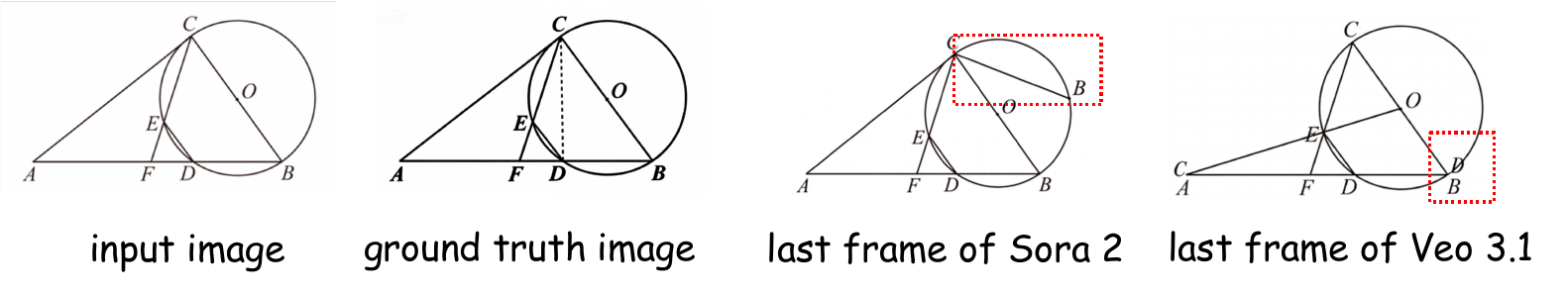}
    \vspace{-0.9cm}
    \caption{\textbf{Showcase of Sora-2's in geometry task}. Both Sora-2 and Veo-3.1 failed to identify the existing point D in the static image. This perceptual error caused them to ignore the instruction connect point C to point D' and instead hallucinate a new point D, connecting C to this incorrect location.}
    \vspace{-0.4cm}
    \label{fig:model_pks_heatmap}
\end{figure}

\noindent{\textbf{Takeaway}}.  Models expose a critical perceptual flaw in complex, symbol-reliant tasks (like geometry): they fail to parse in-context abstract symbols (like "D") as addressable logical components, instead misinterpreting them as incidental visual noise.

\noindent\colorbox{lightgray}{\textbf{Case 4: Algorithmic \& Logical Reasoning (Sudoku)}}

\begin{figure}[H]
    \centering
    \vspace{-0.4cm}
    \includegraphics[width=\linewidth]{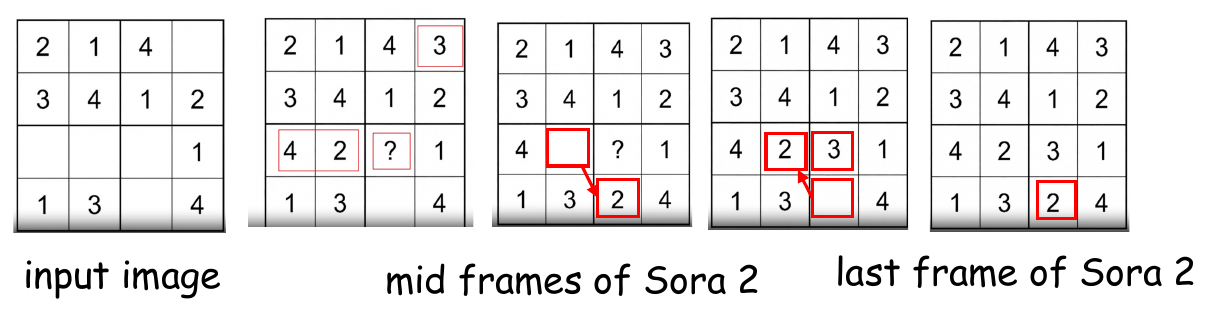}
    \vspace{-0.9cm}
    \caption{\textbf{Showcase of Sora-2's Sudoku task.} In the Sudoku task, Sora-2 exhibits an emergent, human-like thinking process. The model uses a question mark (?) as a placeholder for the unknown value in the third row. This suggests it can hold an internal state of the problem (``this cell is unsolved"). Following the placeholder, the model generates frames that simulate the ``moving" of numbers (2) into their correct, logically-deduced positions.}
    \vspace{-0.4cm}
    \label{fig:model_pks_heatmap}
\end{figure}

\noindent{\textbf{Takeaway}}. Sora-2's Sudoku solving process (using ``?" placeholders and ``moving" numbers) indicates it is acquiring a genuine algorithmic capability, as it is simulating the problem-solving process of ``following Sudoku rules" rather than just pattern-matching the final answer.

\subsubsection{Quantitative Results}

Our evaluation of 7 state-of-the-art models across six reasoning dimensions reveals a clear performance hierarchy, as shown in Figure~\ref{fig:radar_main} and Table~\ref{tab:main_table}. \textbf{Sora-2 achieves the highest overall score (0.560)}, establishing the top tier with particularly strong performance in the most cognitively demanding domains: ``Abstract Reasoning'' (0.604), ``Algorithmic \& Logical'' (0.472), and ``Perceptual'' (0.496). The second tier comprises three highly competitive models—Hailuo-2.3 (0.493), Wan-2.5 (0.490), and Veo-3.1 (0.486)—each exhibiting distinct specialized strengths. \textbf{Hailuo-2.3 achieves the highest score in ``Planning'' (0.778)}, showcasing exceptional sequential decision-making capabilities, while \textbf{Wan-2.5 leads in ``Analogy'' (0.500)}, excelling at analogical reasoning. Veo-3.1 has balanced performance, ranking second in both ``Algorithmic \& Logical'' (0.451) and ``Planning'' (0.722). In contrast, Kling-v1 (0.198) and Seedance-1.0-Lite (0.279) form the lower tier with scores substantially below the leading models, indicating considerable room for improvement in reasoning capabilities.

\section{Conclusion}
\vspace{-0.08cm}
Video generation models are transitioning from visual synthesizers to potential world simulators capable of physically-grounded reasoning. However, without rigorous evaluation, we cannot distinguish genuine understanding from sophisticated pattern matching. Gen-ViRe addresses this by providing systematic assessment across six cognitive dimensions, establishing the foundation for quantitative science in generative visual reasoning. Our experiments reveal a critical gap: while current models achieve impressive visual fidelity, they exhibit limitations in sustained logical coherence, physics compliance, and multi-step planning. By diagnosing these specific deficits—whether in perceptual grounding, spatial reasoning, or goal-directed planning—Gen-ViRe provides actionable insights to guide targeted improvements. As AI systems increasingly simulate and interact with physical reality, benchmarks like Gen-ViRe are essential for measuring genuine progress toward intelligent world models that reason about the world, rather than merely rendering plausible pixels.

\clearpage
{
    \small
    \bibliographystyle{ieeenat_fullname}
    \bibliography{main}
}


\end{document}